\documentclass[conference]{IEEEtran}
\IEEEoverridecommandlockouts
\usepackage{cite}
\usepackage{amsmath,amssymb,amsfonts}
\usepackage{algorithmic}
\usepackage{graphicx}
\usepackage{textcomp}
\usepackage{xcolor}
\usepackage{comment}
\usepackage{hyperref}
\usepackage{threeparttable}
\usepackage[font=footnotesize]{caption}
\usepackage{afterpage}
\usepackage{float}
\usepackage{balance}
\usepackage{subcaption} 

\usepackage[moderate,tracking=normal]{savetrees}

\def\BibTeX{{\rm B\kern-.05em{\sc i\kern-.025em b}\kern-.08em
    T\kern-.1667em\lower.7ex\hbox{E}\kern-.125emX}}
\begin{document}

\title{Gazing at Failure: Investigating Human Gaze in Response to Robot Failure in Collaborative Tasks
}

\author{\IEEEauthorblockN{Ramtin Tabatabaei}
\IEEEauthorblockA{\textit{The University of Melbourne} \\
Melbourne, Australia \\
stabatabaeim@student.unimelb.edu.au}
\and

\IEEEauthorblockN{Vassilis Kostakos}
\IEEEauthorblockA{\textit{The University of Melbourne} \\
Melbourne, Australia \\
vassilis.kostakos@unimelb.edu.au}
\and

\IEEEauthorblockN{Wafa Johal}
\IEEEauthorblockA{\textit{The University of Melbourne} \\
Melbourne, Australia \\
wafa.johal@unimelb.edu.au}}


\maketitle

\begin{abstract}
Robots are prone to making errors, which can negatively impact their credibility as teammates during collaborative tasks with human users.
Detecting and recovering from these failures is crucial for maintaining effective level of trust from users. However, robots may fail without being aware of it. One way to detect such failures could be by analysing humans' non-verbal behaviours and reactions to failures. This study investigates how human gaze dynamics can signal a robot's failure and examines how different types of failures affect people's perception of robot. We conducted a user study with 27 participants collaborating with a robotic mobile manipulator to solve tangram puzzles. The robot was programmed to experience two types of failures ---executional and decisional--- occurring either at the beginning or end of the task, with or without acknowledgement of the failure. Our findings reveal that the type and timing of the robot's failure significantly affect participants' gaze behaviour and perception of the robot. Specifically, executional failures led to more gaze shifts and increased focus on the robot, while decisional failures resulted in lower entropy in gaze transitions among areas of interest, particularly when the failure occurred at the end of the task.
These results highlight that gaze can serve as a reliable indicator of robot failures and their types, and could also be used to predict the appropriate recovery actions.

\end{abstract}


\section{Introduction}


As robotics advances, the potential for robots to assist people in various domains, such as manufacturing \cite{sauppe_social_2015, terzioglu_designing_2020}, domestic assistance \cite{babel_step_2022, schneiders_domestic_2021, chatterjee_usage_2024} is becoming increasingly evident.  One significant application of robotics is in human-robot teaming, which involves collaboration between humans and robotic systems working together to perform joint activities \cite{mingyue_ma_human-robot_2018}. In human-robot teaming, robots must behave and communicate effectively to maintain alignment within the team \cite{chakraborti_ai_2017}.
 However, as robots become more integrated into our daily lives, ensuring the reliability of these systems is a pressing concern \cite{honig_understanding_2018, desai_effects_2012}. Robot errors are inevitable, much like human errors, due to the inherent uncertainty of the world and the need to make decisions and act in real-time.
If these failures are not managed appropriately, they can negatively impact task success, human safety, trust, and perceptions of the robot’s intelligence \cite{ schaefer_meta-analysis_2016, salem_would_2015, sebo_i_2019, lei_should_2021}. 
These factors are crucial because the degree to which people trust robots influences their willingness to collaborate with them, which is essential for establishing effective human-robot teams \cite{breazeal_social_2016, rossi_matter_2023, huang_anticipatory_2016}. 
However, trust can fluctuate over time; it tends to increase when robots perform well but might drop rapidly when they inevitably make errors. 
In addition, the type of failure (i.e. at the motion execution or task planning level), might significantly affect trust, and robots that demonstrate awareness of their errors show potential in restoring trust, as the saying goes, ‘a fault confessed is half forgiven.’
By focusing on moments when human interactions deviate from expected patterns, strategies can be identified to make these interactions more robust.

\begin{figure}[t]
  \centering
  \includegraphics[width=\linewidth]{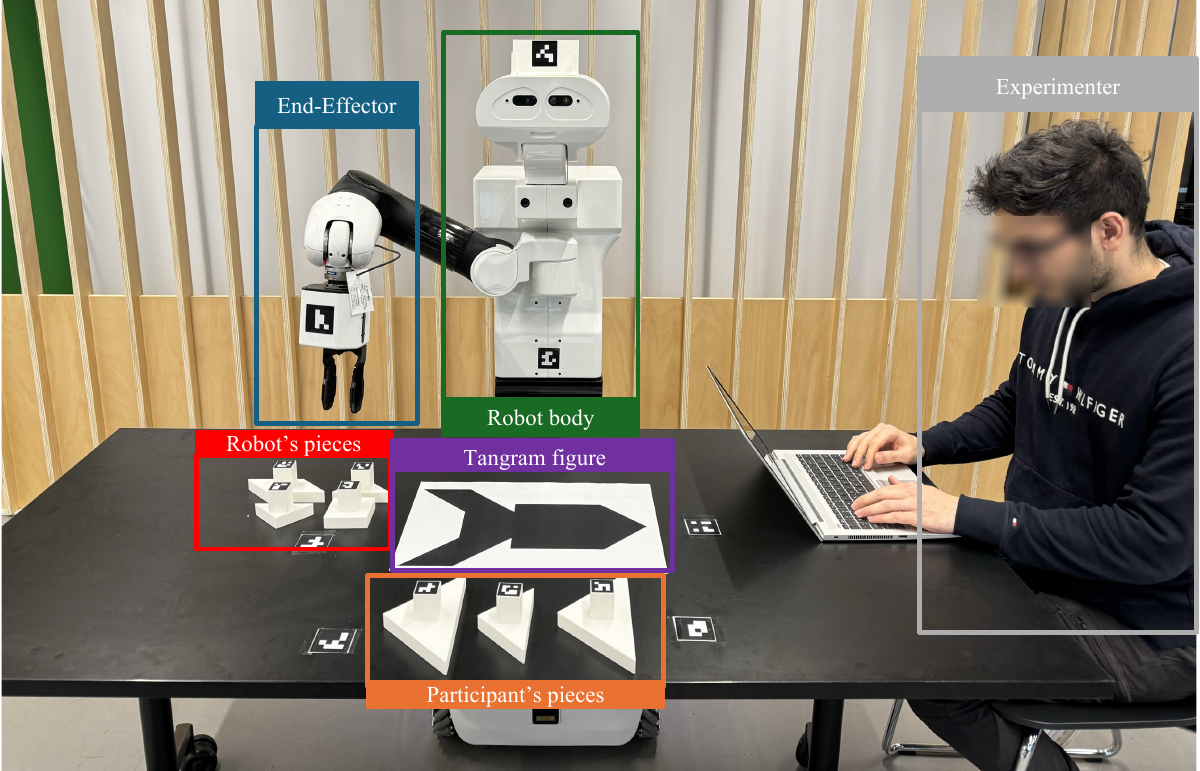} 
  \caption{Different areas of interest in the experiment.
}
  \label{fig:Robot}
\end{figure}

One strategy to enhance human-robot interaction is modelling the user's reactions to robot failures. This user model can be inferred from various signals \cite{ rossi_user_2017}, such as the user’s social cues during the period of the failure \cite{rabinowitz_machine_2018}. One of those social and non-verbal cues is a person’s eye gaze \cite{wachowiak_analysing_2022}, which plays a crucial role in conveying attention \cite{velichkovsky_social_2021, fang_dual_2021}, intentions \cite{velichkovsky_social_2021, fang_dual_2021, rubies_enhancing_2022}, and emotional states \cite{velichkovsky_social_2021, huang_using_2015}. Eye gaze has been leveraged in human-robot interactions to enhance the robot's ability to comprehend and to anticipate human actions \cite{fang_dual_2021, mwangi_dyadic_2018}. Research has also shown that people exhibit consistent gaze patterns while performing specific activities, \cite{johansson_eye-hand_2001, land_what_2001, matthis_gaze_2018}, making it possible to model these patterns. However, there is a lack of studies on accurate gaze patterns in response to robot failures, which could potentially aid robots in recovering from their failures.

This study examines the impact of robot failures on users' perception and gaze behaviour. Utilising a within-between experimental design, we analysed the effects of failure types, failure timing, and failure acknowledgement in a sample of 27 participants. Participants engaged in a collaborative task involving four Tangram puzzles, during which the robot was programmed to fail. Each participant experienced all combinations of failure type and timing (within-subjects), with one group exposed to the robot acknowledging its failures and the other group receiving no acknowledgement (between-subjects). Our results indicate that users' gaze patterns during failure events differ significantly from those observed during times of no failure. Furthermore, these gaze behaviours are highly dependent on the type of failure.

\section{RELATED WORKS}

\subsection{Social Signals to Robot's Failure}

 Social signals have been found to be reliable indicators of errors, as people react to robot errors socially due to their unexpectedness. Specifically, users display more social signals during situations with errors than those without \cite{stiber_using_2023}. Common instinctive responses to robot errors include gaze \cite{kontogiorgos_embodiment_2020, peacock_gaze_2022, kontogiorgos_systematic_2021}, facial expressions \cite{mirnig_impact_2015, stiber_modeling_2022, kontogiorgos_systematic_2021, wachowiak_time_2024}, verbalization \cite{kontogiorgos_embodiment_2020, mirnig_impact_2015, kontogiorgos_systematic_2021}, and body movements \cite{mirnig_impact_2015, trung_head_2017, wachowiak_time_2024}.

Several studies have demonstrated that participants exhibit distinct social signals in response to robot failures. For instance, Aronson et al. \cite{aronson_gaze_2018} observed that participants' gaze patterns deviate from the norm during unexpected robot actions. Wachowiak et al. \cite{wachowiak_analysing_2022} further found that during failures, participants focus more on the entity they are collaborating with, whereas in error-free scenarios, their gaze is more evenly distributed. Similarly, Peacock et al. \cite{peacock_gaze_2022} noted that gaze initially increases in motion during failures and then stabilizes as users recognize and correct the error. Stiber et al. \cite{stiber_using_2023} identified that specific facial muscles, such as those involved in smiling and brow lowering, become more active during robot errors. Kontogiorgos et al. \cite{kontogiorgos_systematic_2021, kontogiorgos_embodiment_2020} found that robot failures lead to increases in spoken words, utterance duration, and gaze shifts towards the robot, indicating heightened engagement during errors. While there is substantial research on human reactions to interacting with a failing robot, there is limited understanding of how this interaction affects the perception of the robot as a teammate in highly collaborative tasks and its impact on human gaze behaviour. This gap is significant, as existing research indicates that individuals exhibit distinct social reactions depending on the type of robot failure encountered \cite{mirnig_impact_2015, kontogiorgos_systematic_2021}.

\subsection{Types of Failure}

 Robot failures can be classified into different types depending on the nature of the issue. Mirnig et al. \cite{mirnig_impact_2015} identified two main types: technical failures, where the robot fails to perform its task correctly, and social norm violations, which occur when the robot deviates from expected social behaviour. 
Honig and Oron-Gilad \cite{honig_understanding_2018} offered a taxonomy distinguishing between (a) technical failures, involving hardware malfunctions and software issues, and (b) interaction failures, arising from uncertainties during interactions with the environment, other agents, or humans. Similarly, Tian et al. \cite{tian_taxonomy_2021} categorised errors into performance errors, which affect perceived intelligence and task competence, and social errors, which impact socio-affective competence. Kontogiorgos et al. \cite{kontogiorgos_embodiment_2020} further classified conversational failures into task-oriented failures, such as incorrect guidance or incomplete instructions, and social protocol violations, like disengagement. Additionally, Morales et al. \cite{morales_interaction_2019} categorised robot failures into Personal Risk Failures (e.g., throwing objects or erratic movements), Property Risk Failures (e.g., dropping or crushing objects), and an Assistance scenario where the robot seeks participant help without posing direct risks.

\subsection{Timing of Failure}
Research has shown that the timing of failures during a task influences people's perceptions of the robot in various ways. Desai et al. \cite{desai_impact_2013} found that early failures significantly reduce trust and make it harder to recover compared to failures occurring later in the interaction. Similarly, Rossi et al. \cite{rossi_how_2017} observed that participants' trust in the robot did not increase when severe mistakes happened early in the interaction. In contrast, Morales et al. \cite{morales_interaction_2019} discovered that the order of failures significantly impacts participants' perceptions, with severe failures occurring last leaving a stronger impression and making participants more likely to believe the robot will fail again in future tasks. Lucas et al. \cite{lucas_getting_2018} also found that early errors can be somewhat recovered from, especially with positive social interaction, but late errors are more damaging. On the other hand, Kontogiorgos et al. \cite{kontogiorgos_systematic_2021} demonstrated that reactions to failures remain consistent, regardless of whether they occur early or late in the interaction. Existing research shows contrasting results regarding the effects of robot failure timing on trust, highlighting the need for further study to understand how failure timing affects user perception of the robot. Furthermore, to the best of the author's knowledge, no research has explored the impacts of failure timing on human gaze behaviour.

\subsection{Failure Repair}
Previous studies have investigated how different trust repair strategies used by robots influence users' perceptions. For example, LeMasurier et al. \cite{lemasurier_reactive_2024} considered three strategies for explaining failures: 1) The robot only acknowledges its failure, 2) The robot explains what went wrong and why after the failure, and 3) The robot predicts and explains potential failures before they occur. Their results highlight that both explaining and predicting failures enhance users' perceptions of a robot's intelligence and trustworthiness compared to providing no explanation at all.
In \cite{esterwood_you_2021}, four trust repair approaches (promises, denials, explanations, apologies) were compared during a collaborative robot task. Apologies, explanations, and promises were similarly effective and outperformed denials for the ability measure, while apologies and promises were most effective for benevolence.
Additionally, Wachowiak et al. \cite{wachowiak_when_2024} found that participants preferred apologies most and silence least when a robot made an error.
While previous studies have shown that apologies and explanations for failures help people regain trust in the robot, we wonder if this could also affect their social behaviour, specifically their gaze behaviour.


\subsection{Gaze in HRC}
The gaze behaviour in human-robot collaboration has been studied widely, focusing on both the robot’s gaze behaviour while collaborating with a human and the human’s gaze behaviour while interacting with a robot \cite{srinivasan_survey_2011, admoni_social_2017}. Most studies in human-robot collaboration emphasise the human’s gaze behaviour, as it can indicate human intent and focus, allowing the robot to determine its next move and adapt its behaviour accordingly.  

The literature collectively highlights the significant role of gaze in intent recognition. Huang et al. \cite{huang_anticipatory_2016} focused on enabling robots to proactively perform task actions by predicting the task intent of their human partners based on observed gaze patterns. Their anticipatory control method significantly improved task efficiency, allowing the robot to respond faster compared to the reactive method. 
Additionally, Shi et al. \cite{shi_gazeemd_2021} also developed an effective model for accurately determining which object a person intends to focus on during interactions with a robot. The literature demonstrates that gaze can be a reliable indicator of a person's intent and by extension their anticipation of upcoming actions. This leads us to wonder whether gaze also has the potential to help the robot repair from its failure.

Despite extensive research on human reactions to robot failures, little is known about how such failures influence perceptions of the robot as a teammate or affect human gaze behavior. Moreover, the impact of failure timing and the robot's acknowledgment on user gaze remains unclear. To explore these gaps, we address the following research questions:

\vspace{0.5em} 

\begin{itemize}
\item \textbf{RQ1} How does human gaze behaviour change in response to different robot failures during a collaborative task?
\vspace{0.5em} 

\item \textbf{RQ2} How do different robotic failures affect human perception of the robot as a teammate?
\end{itemize}

\section{Methodology}

\subsection{Tasks Description}

The experiment consists of four distinct tasks, in which one participant and a robot collaboratively solve Tangram puzzles. In each task, participants were required to create a unique shape using Tangram pieces. The sequence of shapes to solve is Rocket, Rabbit, Turtle, and Cat. We chose puzzles of similar difficulty to ensure the difficulty would not affect participants’ perception and behaviour towards the robot.

Each Tangram puzzle consisted of seven pieces. The robot handled four pieces (two small triangles, a square, and a parallelogram), while the participant had to place correctly three pieces (a medium-sized triangle and two large triangles). The Tangram pieces were 3D printed, and when assembled, formed a square of $200\,\text{mm}$ in side. Each piece was $20\,\text{mm}$ high. Besides, an cube ($32\,\text{mm} \times 32\,\text{mm} \times 40\,\text{mm}$) was attached on top of each piece to act as a handle and to facilitate the robot's ability to pick up the pieces. The puzzles' silhouettes were printed in black on A2 white paper, slightly larger than the Tangram pieces to avoid the need for very precise placement, with approximately 1 cm clearance on each side. These papers were fixed to the table, and the participant and the robot had to place each of their pieces in the correct position. The participant was asked to move a piece only after the robot had completed its action.

The robot always placed the first piece. To reduce confusion about when the participant should place their piece, the robot said: ``Now it is your turn.'', after placing each piece, except for the last one, when it said: ``Now, let’s solve the next puzzle.'' If the participant placed a piece incorrectly, the robot responded,``You have placed the object in the wrong location.''

The robot's pieces were placed next to the paper and near the robot, as shown in Figure \ref{fig:Robot}. In each puzzle, the arrangement of the pieces varied from the previous one, and the robot first determined the placement and orientation of each piece before picking it up. To facilitate this process, an ArUco marker was attached to the top of each piece, allowing the robot to accurately locate them. The Tiago robot, programmed using ROS1, then utilised the tf library to transform the pose of the desired object to the coordinate frame associated with its arm, and subsequently employed inverse kinematics to move its arm to the correct location. The robot's head movements were pre-programmed to approximately mimic human gaze behaviour. During its turn, the robot maintained its gaze on the Tangram piece while picking it up and placing it. Once the robot finished placing a piece, it started looking at the participant.

\subsection{Robot Failures}
We designed the robot to fail during each task in its interaction with the participants. These failures varied based on their type, timing, and whether the robot acknowledged its failure or not.

\subsubsection{\textbf{Types of Failures}}

The types of failures in our experiment represent typical robot malfunctions that may occur during interactions and are commonly reported in HRI. In this research, the robot will simulate two distinct types of failures: 1) Executional failure (EF) and 2) Decisional failure (DF).

EF can be categorised as a technical failure, specifically timing and ordering \cite{honig_understanding_2018}. In this scenario, the robot pauses for 15 seconds just before picking up an object, while keeping the object within its end effector. After the 15-second pause, the robot will resume and complete the task of picking up and placing the object. This type of failure aligns with previous research \cite{kontogiorgos_embodiment_2020, wachowiak_analysing_2022}.

DF can also categorized as a technical failure, where the robot performs the correct action incorrectly \cite{honig_understanding_2018}. In this scenario, after picking up an object, the robot will mistakenly move to the location designated for a different object, place it and pause for 5 seconds. While still holding the object, the robot will then lift the object again and place it in the correct location. This type of failure aligns with previous research, in which the robot attempts to perform the correct action but executes it incorrectly \cite{inceoglu_fino-net_2021, stiber_using_2023}.

The procedure for picking up and placing objects during failure events is identical to the procedure when no failure occurs, indicating that the robot shows no signs of committing a failure beforehand. The only distinction in the EF is a pause, which increases the total time for the pick-and-place task by 15 seconds. In the DF, the robot moves its arm to the wrong location, goes down, and comes back up, resulting in an overall increase of 16.5 seconds to the motion.

\subsubsection{\textbf{Timing of Failures}}

The literature suggests that the timing of a failure—whether it occurs at the beginning or the end of an interaction—can affect a person’s perception of the robot differently. 
In this research, we aim to investigate how the timing of a failure impacts both gaze behaviour and user perceptions. Specifically, the robot may fail either at the beginning of the collaboration when placing its first piece, or towards the end of the interaction when placing its third piece.
 
\subsubsection{\textbf{Acknowledgement of Failures}}
A fault confessed is half redressed. Guided by this principle, we explored how the robot’s ability to acknowledge its mistakes influences participants’ perception and gaze patterns in subsequent failures. We designed two distinct scenarios. In one scenario, the robot demonstrates awareness of its mistakes by acknowledging each failure immediately after they occur. After a DF failure, it says, ``Sorry, I made a mistake.'' and after an EF, it says, ``Sorry for the delay.'' In the other scenario, the robot does not declare any of its failures.  In both scenarios, the robot performs physical repairs.

\subsection{Participants}

\begin{figure*}[t!]
  \centering
  \includegraphics[width=0.8\linewidth]{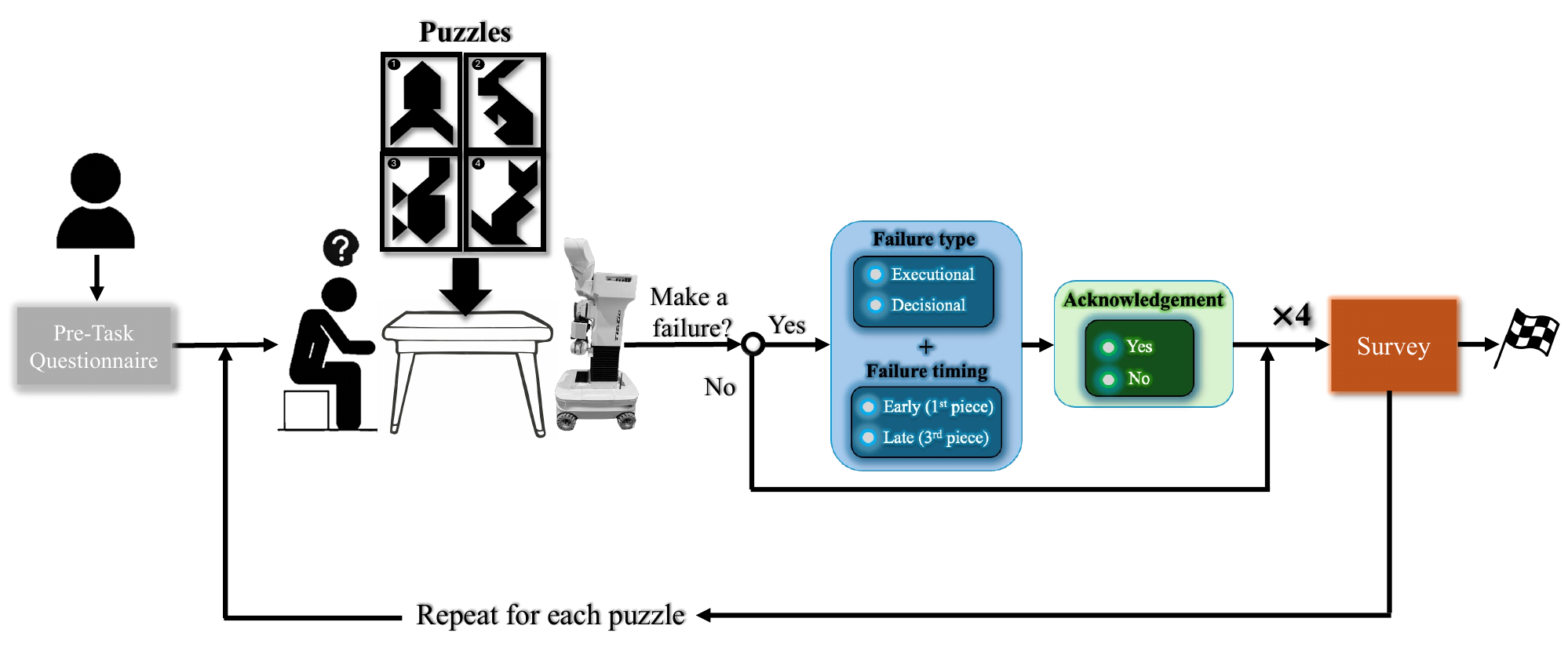} 
  \caption{Experimental diagram showing the process where participants first complete a pre-task questionnaire, followed by collaboratively solving a Tangram puzzle four times.}
  \label{fig:Experimentdiagram}
\end{figure*}

We conducted \textit{a priori} power analysis to calculate the sample size for our experiment using \textit{G*Power}~\cite{faul_gpower_2007}. The calculation was based on a medium effect size of $f=0.25$, an alpha level of 0.05, and a power of 0.8. As a result, we determined a minimum of 24 participants was required; however, we recruited 27 participants (16 females, 10 males, and 1 non-binary) via our university recruitment website. Participants primarily consisted of students and university staff, none of whom had previous experience working with robots. They were compensated with a gift voucher for their participation. The average age was ranging between 18y.o. and 34y.o.  (M = 23.26, SD = 4.3). Due to some technical issue, one participant's gaze data was not recorded, and another participant did not complete some questions in one of the after-task questionnaires. Participants signed a consent form before participation and were made aware that their gaze data was recorded during the experiment. At the end of the experiment, participants were informed that the study aimed to observe their responses to the robot's failures.
\subsection{Experiment}

 Participants were first briefed by an experimenter on how to interact with the robot and the goals of the tasks. They then completed a demographics questionnaire, providing their age and gender. Following this, participants were seated at a table opposite the robot and asked to wear eye-tracking glasses during the experiment. As per the experimental conditions (Table \ref{table:participant_data}), during each puzzle, the robot correctly picked and placed three pieces but intentionally made an error with one piece. After each puzzle, participants completed a questionnaire assessing their perception of the robot's performance during that specific puzzle. Participants were unaware that these errors were pre-programmed. This cycle was repeated for all four puzzles. The experimental procedure is illustrated in Figure \ref{fig:Experimentdiagram}.

The experimenter initiated the robot's turns and intervened when participants made mistakes by triggering the robot's verbal response. This ensured that the robot began its turn immediately after the participant's turn, maintaining a consistent time gap across all participants. The experimenter was seated on the opposite side of the table, near the robot, to ensure safety and to press the emergency button in case of an actual malfunction. For consistency, the same experimenter conducted all sessions and operated the robot throughout the study.

The experiment was conducted in a laboratory on the University of Melbourne campus. The duration of solving each puzzle together with the robot was about 191.12s ± 35.40s. After each puzzle, the experimenter asked the participant to complete a survey and prepared the table for the next puzzle. The gap between each puzzle was about 93.88s ± 38.81s.

A mixed experimental design was used, with failure types (executional and decisional) and failure timings (early and late) as within-subjects factors, and failure acknowledgement as a between-subjects factor. To minimise order effects, the within-subjects factors were counterbalanced using a four-condition balanced Latin Square. Each factor was systematically integrated into the puzzles. The first thirteen participants experienced the failure acknowledgement, while the second fourteen did not.

\begin{table}[h!]
\centering
\resizebox{0.48\textwidth}{!}{
\begin{tabular}{ c|c c c c c}

\textbf{Participant ID}  & \textbf{Puzzle 1} & \textbf{Puzzle 2} & \textbf{Puzzle 3} & \textbf{Puzzle 4} & \textbf{Acknowledgment} \\ \hline
1  & EF (Early) & EF (Late) & DF (Late) & DF (Early) & Yes \\ 
2  & EF (Late) & DF (Early) & EF (Early) & DF (Late) & Yes\\ 
3  & DF (Early) & DF (Late) & EF (Late) & EF (Early) & Yes\\ 
4  & DF (Late) & EF (Early) & DF (Early) & EF (Late) & Yes\\
5  & EF (Early) & EF (Late) & DF (Late) & DF (Early) & Yes \\
...  &  ... &  ... & ... & ... &  ... \\ 
14  & EF (Early) & EF (Late) & DF (Late) & DF (Early) & No \\ 
...  &  ... &  ... & ... & ... &  ... \\
\end{tabular}

}
\caption{Order of failure type and timing across puzzles with acknowledgement of failure}
\label{table:participant_data}
\end{table}

\subsection{Measures}
\subsubsection{Objective Gaze Measures}\label{gaze_measures}

For each puzzle and each piece, we recorded the robot’s current action— such as moving above the target object, and lowering to pick up the object—along with whether a failure occurred and the type of failure, all based on Unix time. We recorded users’ gaze data during the whole experiment.

Gaze data was collected during the tasks as participants collaborated with the Tiago robot to solve the puzzles. In our experiment, the gaze data during the robot's turn was particularly important, from the moment it started moving until it completed its turn. Data was captured using Neon Eye Tracking Glasses from Pupil Labs. The gaze data included the participant’s field of view image frame along with the x and y coordinates of their gaze within that frame. This data was recorded in real-time on a computer. The gaze data was captured at a rate of 30 Hz for both the image frames and gaze coordinates.

To facilitate the identification of participants' areas of interest (AoIs), we attached ArUco markers near the areas of interest. The AoIs in our experiment included the robot body (comprising the robot's face and torso), the Tangram figure, the end effector, the robot’s pieces, the participant’s pieces, and the experimenter. These areas of interest are illustrated in Figure \ref{fig:Robot}.



We calculated several gaze-related measures to analyse user behaviour during the interaction. These metrics included: (1) the number of gaze shifts toward the robot body, (2) the number of gaze shifts across all AoIs, (3) the proportional distribution of gaze directed toward the robot body, the Tangram figure, and the robot’s end effector, and (4) transition and stationary entropy derived from gaze transition matrices \cite{krejtz_gaze_2015, ebeid_analyzing_2019}. Each of these measures captures different aspects of gaze behaviour. The number of gaze shifts reflects the frequency of visual transitions between specific areas, providing insight into user engagement and focus dynamics. The proportional distribution of gaze indicates how much time users spent looking at each AoI, offering a measure of relative visual attention. Transition entropy quantifies the unpredictability of gaze transitions between AoIs, while stationary entropy measures the overall distribution of gaze within the AoIs, highlighting how scattered or concentrated the gaze behaviour was during the task.

The gaze measures were calculated during a specific time window for both failure and non-failure conditions: from the moment the robot began moving to pick up an object until it placed the object and returned to its initial position.
Since failure timing is not applicable in non-failure conditions, the analysis of these measures was conducted in two ways. First, we analysed the data by failure type (no failure, executional failure, decisional failure) and acknowledgement (yes vs. no). Second, we analysed it by failure type (executional failure, decisional failure), timing (early vs. late), and acknowledgement (yes vs. no).

\subsubsection{Subjective Measures}\label{Subjective_Measures}

After each puzzle, participants rated their perceptions of the robot's behaviour in terms of perceived intelligence, perceived safety, and performance trust. Perceived intelligence and safety were measured using items from the Godspeed questionnaire \cite{bartneck_measurement_2009}, while performance trust was assessed using items from the Multi-Dimensional Measure of Trust (MDMT) questionnaire \cite{ullman_mdmt_2023}.

To evaluate the level of intelligence participants attributed to the robot, we used three items from the Godspeed questionnaire: “Incompetent/Competent,” “Irresponsible/Responsible,” and “Foolish/Sensible.” For perceived safety, we included one item from the Godspeed questionnaire: “Anxious/Relaxed.” To assess performance trust across various robot failures, we utilised the “performance trust” dimension from the MDMT. This included two items from the Reliable subscale (“Reliable” and “Predictable”) and two items from the Competent subscale (“Skilled” and “Capable”).

The analysis of these measures was conducted based on failure type (executional failure, decisional failure), timing (early vs. late), and acknowledgement (yes vs. no).

\section{RESULTS}



\subsection{Behavioural Response}

In this section, we address the first research question by analysing participants' gaze behaviour using the measures outlined in section \ref{gaze_measures}. Our analysis focuses on how these metrics vary during failure situations. Additionally,  we investigate the anticipatory capability of participants' gaze about the placement of objects.


\subsubsection{Gaze Shift}

\begin{figure}[t]
  \centering
  \begin{subfigure}[t]{0.48\linewidth}
    \centering
    \includegraphics[width=\linewidth]{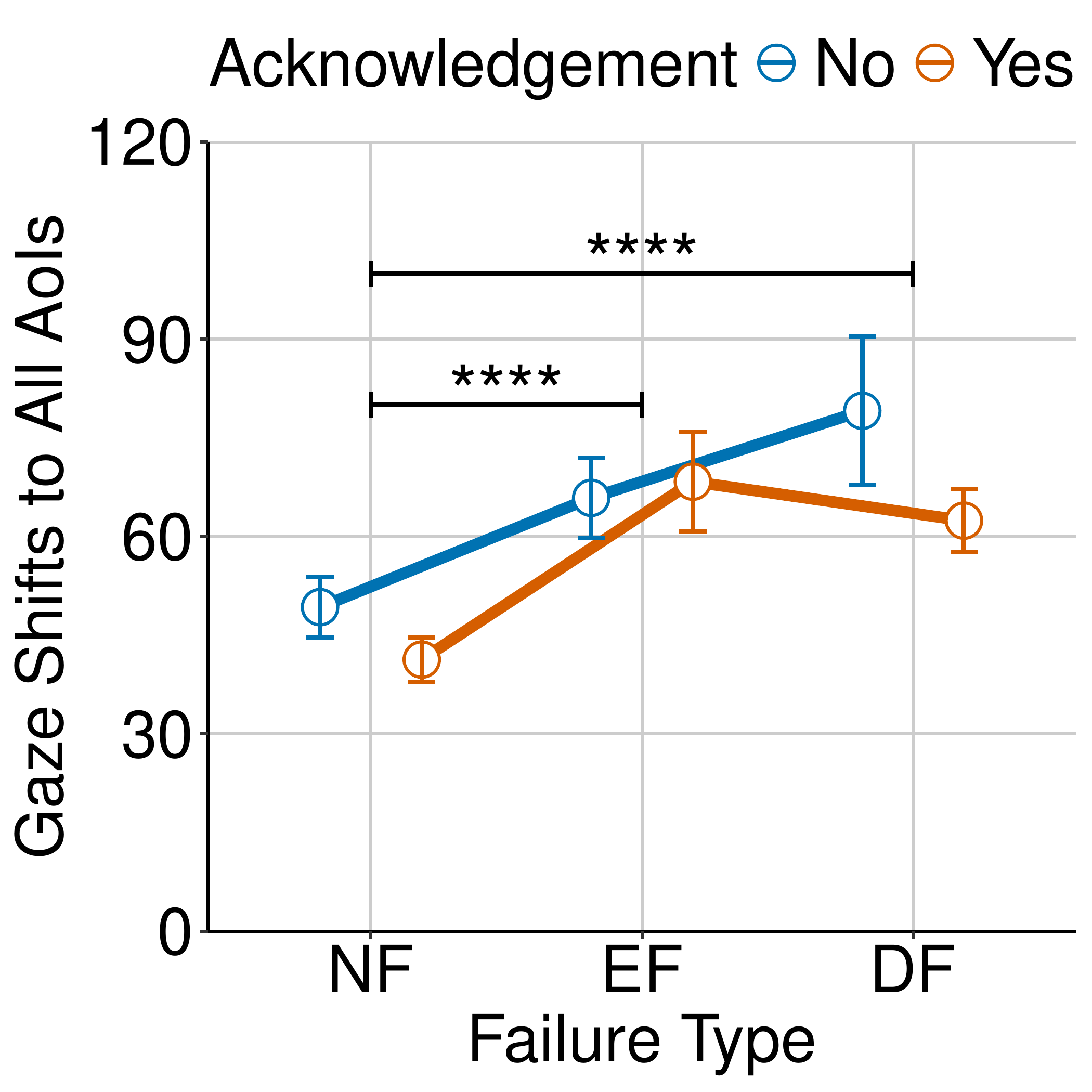}

    \label{GazeShiftsAllAoIs}
  \end{subfigure}
  \hfill
  \begin{subfigure}[t]{0.48\linewidth}
    \centering
    \includegraphics[width=\linewidth]{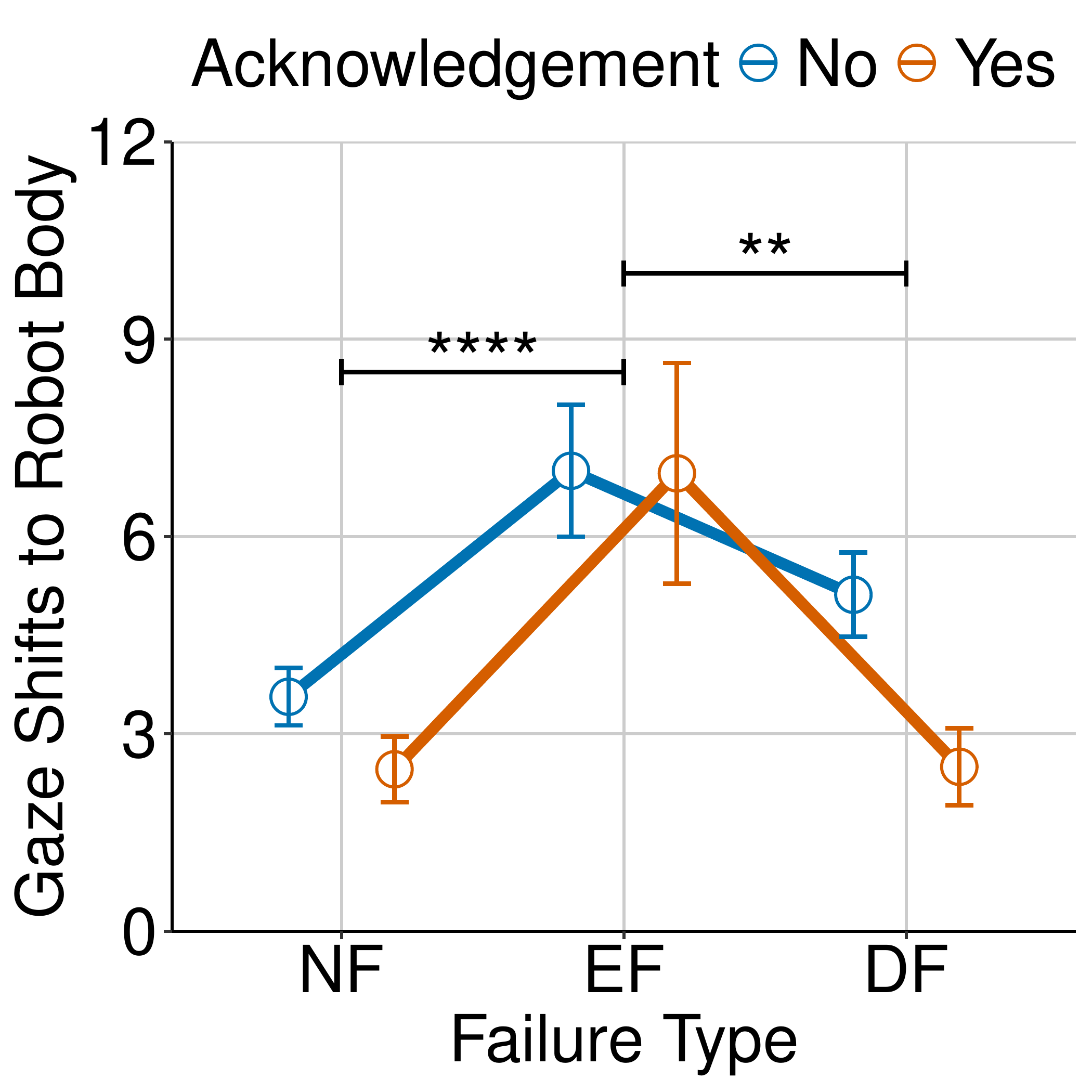}

    \label{GazeShiftsRobotBody}
  \end{subfigure}
  \caption{The average number of gaze shifts across all AoIs (left) and toward the robot body (right) across three different failure situations, with or without the robot acknowledging its failure. Error bars represent the standard error of the mean. Significance levels, based on adjusted p-values, are denoted as follows: ** for $p < .01$, and **** for $p < .0001$.}
  \label{GazeShifts2Conditions}
\end{figure}

 We conducted a two-way ANOVA to compare gaze patterns during failure versus non-failure robot actions, with having failure type as a within-subjects and the acknowledgement as a between-subjects. For the number of gaze shifts across all AoIs, the results indicated a significant main effect for the factor of failure type ($F(2,48)=15.16$; $p<.001$; $\eta^2=0.39$). Bonferroni-corrected pairwise t-tests revealed significant differences between each type of failure and no failure (NF). For the number of gaze shifts toward the robot’s body (i.e., the robot’s face and torso), the results again showed significant main effects for the factor of failure type ($F(2,48)=21.48$; $p<.001$; $\eta^2=0.47$). Bonferroni-corrected pairwise t-tests indicated significant differences between EF and DF, as well as between EF and NF. Figure \ref{GazeShifts2Conditions} illustrates the average values for each condition.

\begin{figure}[t]
  \centering
  \begin{subfigure}[t]{0.48\linewidth}
    \centering
    \includegraphics[width=\linewidth]{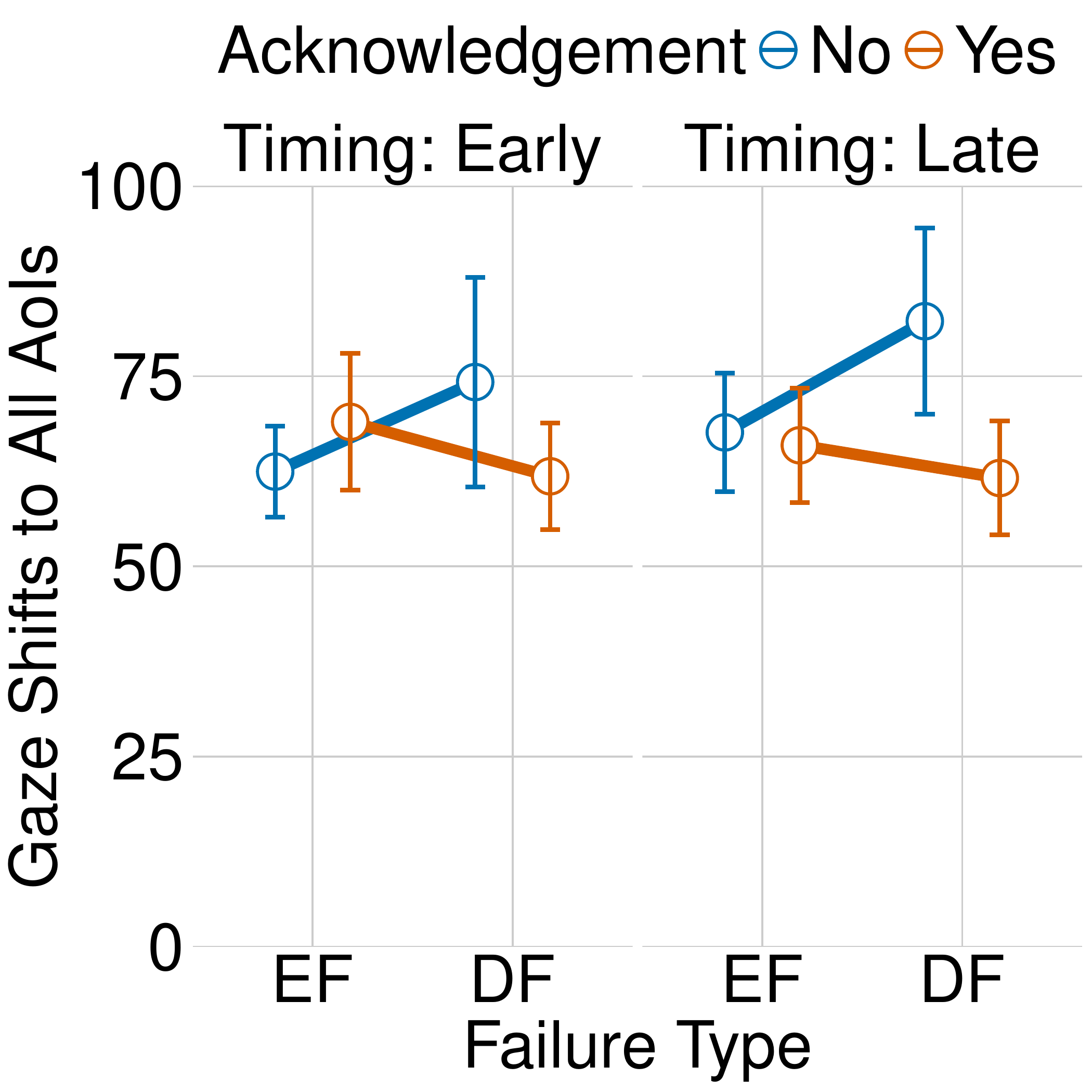}
  \end{subfigure}
  \hfill
  \begin{subfigure}[t]{0.48\linewidth}
    \centering
    \includegraphics[width=\linewidth]{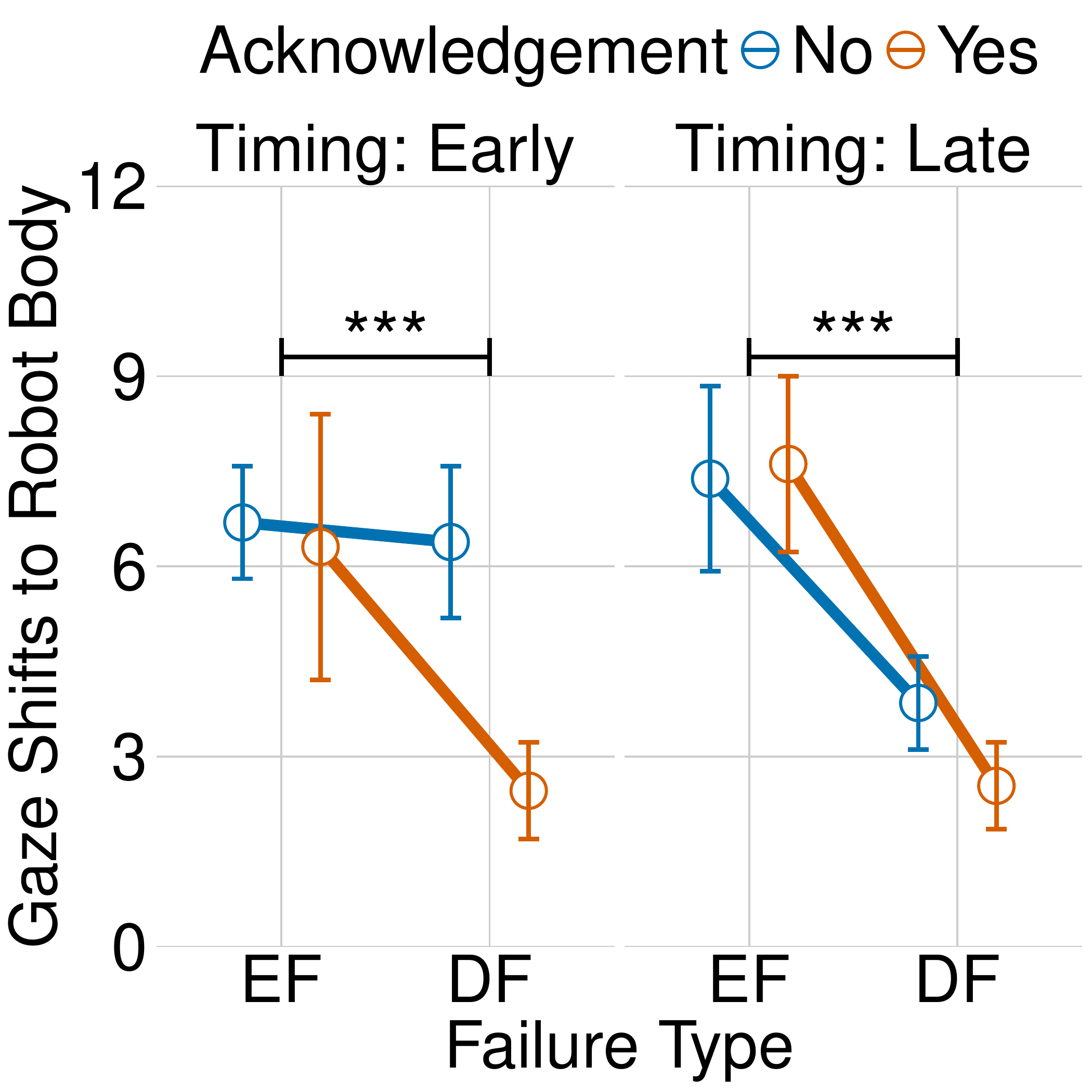}
  \end{subfigure}
  \caption{The average number of gaze shifts across all AoIs and the average number of gaze shifts toward the robot body, comparing failure type, failure timing, and the robot’s acknowledgement of its failure. Error bars represent the standard error of the mean. Significance levels are denoted as follows: *** for $p < .001$.}
  \label{GazeShifts3Conditions}
\end{figure}

Subsequently, a three-way ANOVA was conducted to analyse gaze patterns during failure durations, focusing on the effects of failure type and timing as within-subjects factors, and acknowledgement as a between-subjects factor. For the number of gaze shifts across all AoIs, the results showed no significant effects for any of the factors. However, for the number of gaze shifts toward the robot’s body, the results revealed a significant effect of failure type ($F(1,24)=17.780$; $p<.001$; $\eta^2=0.43$). Figure \ref{GazeShifts3Conditions} shows the average values for each condition.


\subsubsection{Gaze Distribution}

In this section, we compare the proportion of gaze directed toward three AoIs: the robot’s end effector, the Tangram figure, and the robot’s body, during each task while the robot is performing its actions.

\begin{figure}[t!]
  \centering
  \includegraphics[width=\linewidth]{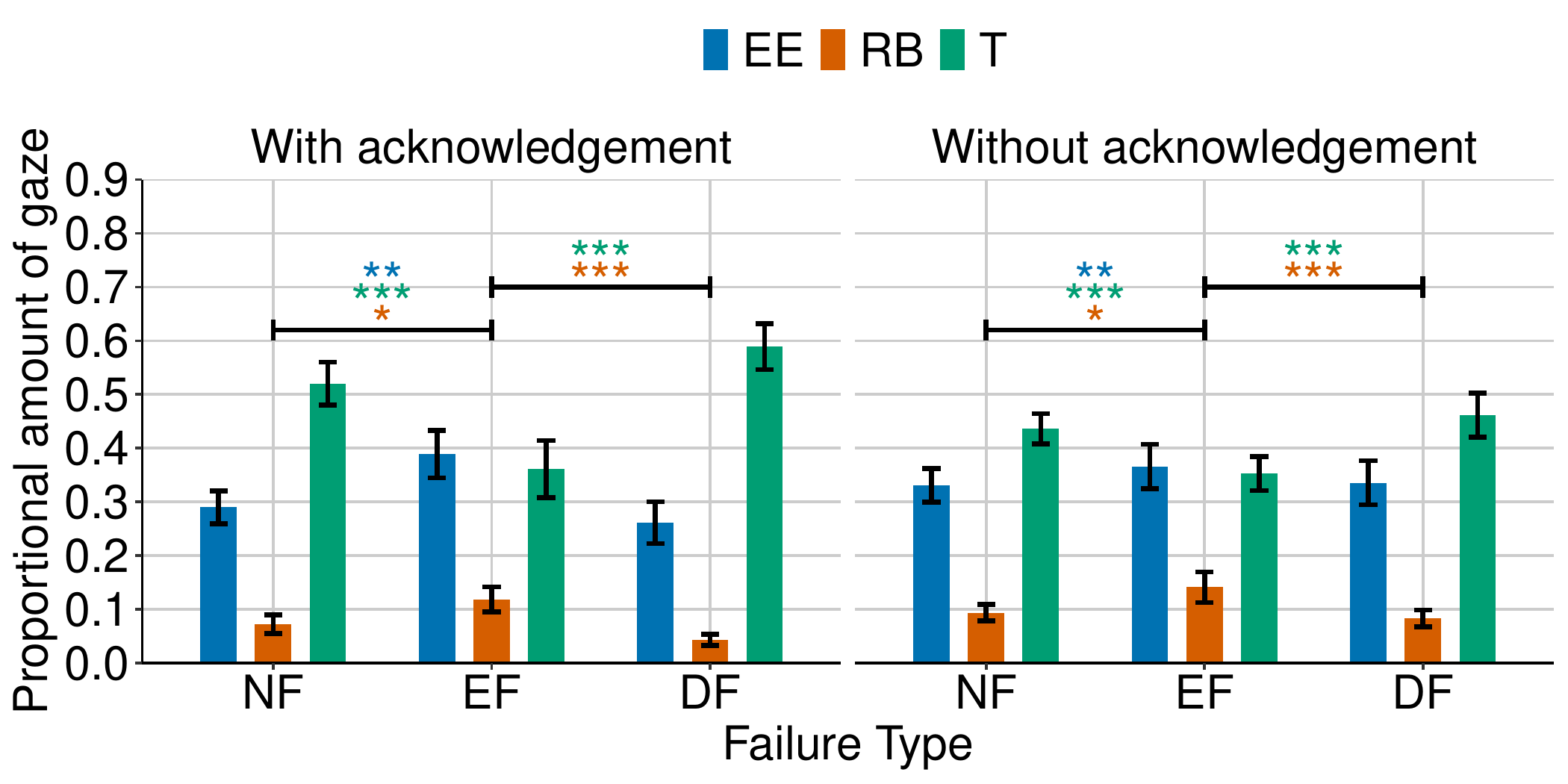} 
  \caption{The average proportion of participant gazes directed at the end effector (EE), robot body/face (RB), and Tangram figure (T) during puzzle solving, across three different failure situations, with or without the robot acknowledging its failure. Error bars represent the standard error of the mean. Significance levels, based on adjusted p-values, are denoted as follows: * for $p < .05$, ** for $p < .01$, and *** for $p < .001$.}
  \label{proportions2conditions}
\end{figure}

First, we compare the proportion of gaze directed during failure events to that during non-failure events. The results of the two-way ANOVA revealed significant differences in failure type but no differences in acknowledgement across all measures. Specifically, significant differences were observed for the end effector ($F(2,48)=6.13$; $p = .009$; $\eta^2=0.20$), the Tangram figure ($F(2,48)=17.71$; $p<.001$; $\eta^2=0.42$), and the robot's body ($F(2,48)=14.35$; $p<.001$; $\eta^2=0.37$). Bonferroni-corrected pairwise tests indicated significant differences between EF and NF, as well as between EF and DF for all measures. Figure \ref{proportions2conditions} shows the average values for each measure.

We subsequently conducted a three-way ANOVA with failure type and timing as within-subjects factors, and acknowledgement as a between-subjects factor. The results, as presented in Table \ref{gazedis}, demonstrated significant differences in failure type and timing across the end effector, Tangram figure, and Robot body. Notably, for the Robot body, we also observed significant interactions between acknowledgement and timing.
Our analysis showed that participants looked at the Tangram figure more when the failure occurred early in the interaction compared to late failures, while they focused more on the robot's body and end effector during late failures than early ones.

\begin{table*}[tbh!]
\centering
\resizebox{0.9\textwidth}{!}{
\begin{tabular}{c||c||c c c c c c c}

Scale & Measure & Type & Timing & Acknowledgement & [Type*Timing] & [Type*Acknowledgement] & [Timing*Acknowledgement] & [Type*Timing*Acknowledgement] \\ \hline
End Effector & df & $$(1,24)$$ & $(1,24)$ & $(1,24)$ & $(1,24)$ & $(1,24)$ & $(1,24)$ & $(1,24)$ \\ 
 & F value & 7.79 & 4.49 & 0.31 & $<$0.001 & 3.85 & 0.55 & 0.75 \\
& p value & \textbf{.010} & \textbf{.447} & .586 & .980 & .061 & .465 & .396 \\ 
& $\eta^2$  & 0.24 & 0.16 & 0.01 & $<$0.0001 & 0.14 & 0.02 & 0.03 \\ \hline
Tangram figure & df & $(1,24)$ & $(1,24)$ & $(1,24)$ & $(1,24)$ & $(1,24)$ & $(1,24)$ & $(1,24)$ \\ 
& F value & 25.86 & 4.85 & 2.02 & 0.03 & 4.09 & 0.80 & 2.09 \\
& p value & \textbf{$<$.001} & \textbf{.038} & .168 & .868 & .054 & .379 & .161 \\
& $\eta^2$  & 0.52 & 0.17 & 0.08 & $<$0.01 & 0.15 & 0.03 & 0.08 \\  \hline
Robot Body & df & $(1,24)$ & $(1,24)$ & $(1,24)$ & $(1,24)$ & $(1,24)$ & $(1,24)$ & $(1,24)$ \\ 
 & F value & 23.01 & 5.29 & 1.20 & 2.80 & 0.46 & 5.71 & 1.34 \\
& p value & \textbf{$<$.001} & \textbf{.030} & .284 & .108 & .502 & \textbf{.025} & .258 \\ 
& $\eta^2$  & 0.49 & 0.18 & 0.05 & 0.10 & 0.02 & 0.19 & 0.05 \\ \hline

\end{tabular}
}
\caption{Results of the three-way mixed ANOVA for gaze distribution}
\label{gazedis}
\end{table*}



\subsubsection{Gaze Transition Matrix}

Based on the AoIs, we created transition matrices, focusing exclusively on transitions between different AoIs while excluding self-repeating transitions. We then compared the transition matrices using transition entropy and stationary entropy.

\begin{figure*}[t!]
  \centering
  \includegraphics[width=0.9\linewidth]{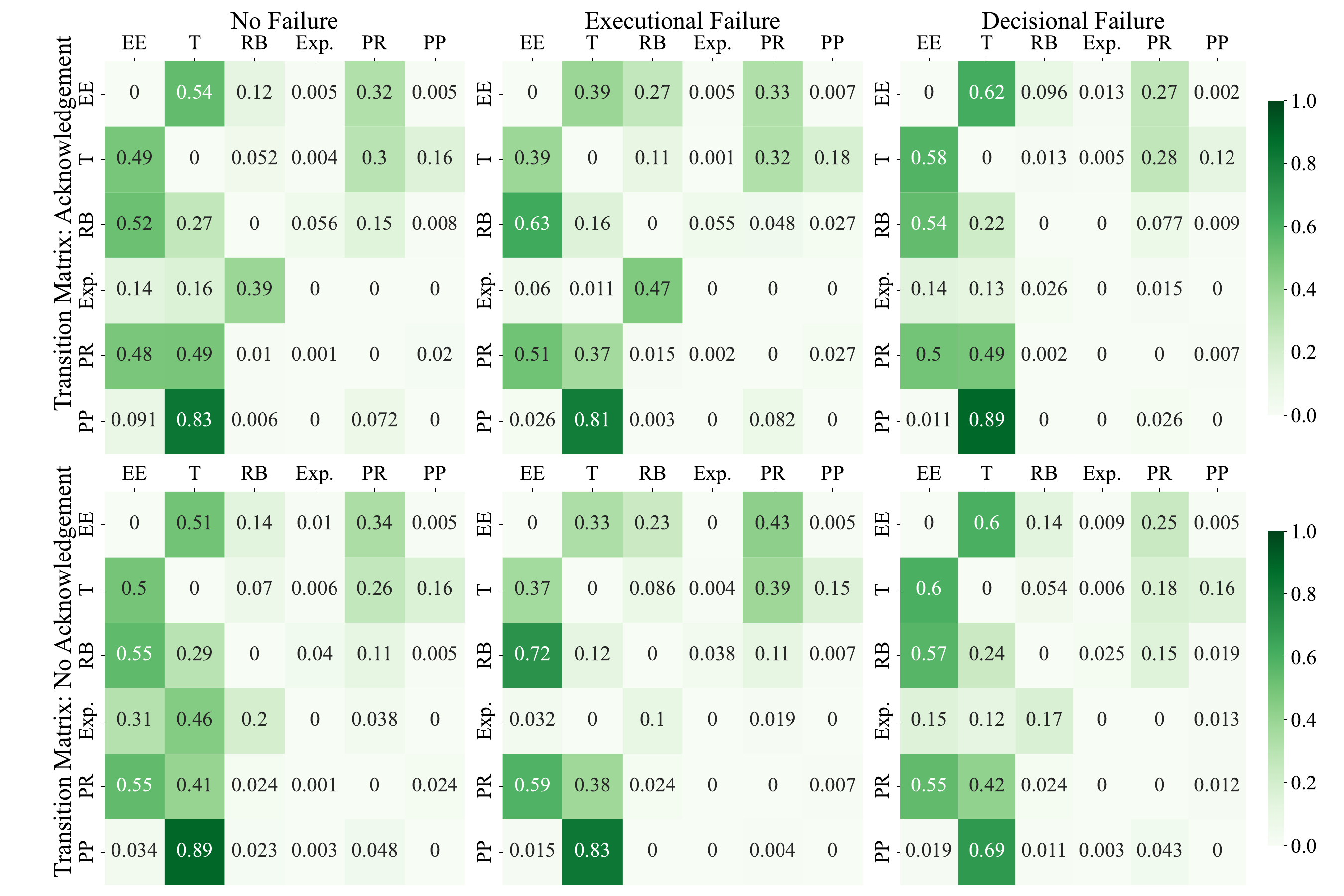} 
  \caption{The transition matrices for three different interaction scenarios—NF, EF, and DF—are presented, both for cases where the robot acknowledges its failure and where it does not. The vertical axis represents the current states, while the horizontal axis represents the next states. 'EE' stands for End Effector, 'T' for Tangram figure, 'RB' for Robot Body, 'Exp.' for Experimenter, 'PR' for Pieces (Robot), and 'PP' for Pieces (Participant). The transition matrices are displayed as heat maps.}
  \label{Transition_matrices}
\end{figure*}

We conducted a two-way ANOVA to compare transition matrices during failure versus non-failure robot actions. The results showed significant differences in failure type for both entropies. Specifically, significant differences were observed for transition entropy ($F(2,48)=13.90$; $p < .001$; $\eta^2=0.37$), and stationary entropy ($F(2,48)=11.01$; $p < .001$; $\eta^2=0.31$). No significant differences were found for acknowledgement. Further, Bonferroni-corrected pairwise t-tests indicated significant differences in transition entropy between EF and NF,
 and between NF and DF.
For stationary entropy, significant differences were observed between EF and DF,
and between NF and DF.
The mean values for transition entropy indicate that NF has the highest value, while DF has the lowest. In contrast, for stationary entropy, EF has the highest value, and DF the lowest. In all conditions where the robot acknowledges its failure, both entropy values are lower.  
The transition matrices are shown in Figure \ref{Transition_matrices}.



Subsequently, we performed the Wilcoxon Signed-Rank Test for failure type and timing, and the Wilcoxon Rank-Sum Test for acknowledgement. The results (Table \ref{TranstionEntropy}) indicate significant differences in failure type for both entropies. Additionally, significant differences in timing were observed for stationary entropy. No significant differences were found for acknowledgement. The median values show that when the failure type is DF, the timing is late, or the robot acknowledges its failure, both entropy values are lower.

\begin{table}[!h]
\centering
\resizebox{0.45\textwidth}{!}{
\begin{tabular}{c||c||c c c}

Scale & Measure & Type & Timing & Acknowledgement \\ \hline
Transition Entropy & N & 52 & 52 & 52 \\ 
 & W & 1007 & 864 & 1583 \\ 
& p-value & \textbf{.004} & .112 & .134  \\ 
 \hline
Stationary Entropy & N & 52 & 52 & 52 \\ 
 & W & 1197 & 1035 & 1522 \\ 
& p-value & \textbf{$<$.001} & \textbf{.002} & .270 \\ 
 \hline

\end{tabular}
}

\caption{Results of the Wilcoxon tests for the entropy of the transition matrices, where N represents the sample size for each condition and W is the test statistic.}
\label{TranstionEntropy}
\end{table}

\subsubsection{Goal Anticipation by Gaze Analysis}

In this section, we examine the proportion of time participants spent looking at the correct goal location for object placement within the Tangram figure, compared to the total time spent looking inside the Tangram figure. The purpose of this analysis is to determine whether participants exhibited anticipatory gaze behaviour to assist the robot in recovering from its failures. Specifically, we assessed the average percentage of time participants looked at the goal, as well as the average number of gaze shifts towards the goal during each puzzle and failure period.
The duration considered for each puzzle spanned from the moment the robot's end effector was positioned above the object it intended to pick up until it was positioned above the designated placement location for that object. During failure periods, we focused on the time from when the robot initiated a failure until it began its repair.

The results indicate that as participants progressed through the puzzles, the proportion of time spent looking at the goal decreased. Specifically, in Puzzle 1, participants looked at the goal for 47\% ($\pm$14\%) of the time, followed by 44\% ($\pm$13\%) in Puzzle 2, 38\% ($\pm$16\%) in Puzzle 3, and 22\% ($\pm$9\%) in Puzzle 4. Additionally, the total number of gaze shifts towards the goal also almost decreased as participants advanced through the puzzles. The average number of gaze shifts per piece was 10.33 ($\pm$5.35) in Puzzle 1, 10.65 ($\pm$4.57) in Puzzle 2, 7.66 ($\pm$4.79) in Puzzle 3, and 5.47 ($\pm$3.18) in Puzzle 4.


When analysing the failure periods, the results show that participants spent 35\% ($\pm$26\%) of their task-related gaze time looking at the goal during EF. This percentage was higher, at 41\% ($\pm$15\%), during DF. Additionally, participants exhibited an average of 3.42 ($\pm3.36$) gaze shifts for each EF, compared to a substantially higher average of 14.84 ($\pm7.11$) gaze shifts for each DF.

\subsection{Subjective Measures}
In this section, we address the second research question by analysing participants' subjective behaviour using the measures outlined in section \ref{Subjective_Measures}. To achieve this, we conducted a three-way ANOVA for each subjective scale to examine the effects of failure type, timing, and acknowledgement. For the Competent scale, significant interaction effects were found for type*timing ($F$(1,24)$ = 6.79$, $p = .016$, $\eta^2 = 0.22$). The Sensible scale showed a significant main effect of timing ($F$(1,24)$ = 5.79$, $p = .024$, $\eta^2 = 0.19$). In the Anxious/Relaxed (Self) scale, significant main effects were observed for timing ($F$(1,24)$ = 7.80$, $p = .010$, $\eta^2 = 0.24$) and acknowledgement ($F$(1,24)$ = 5.50$, $p = .027$, $\eta^2 = 0.18$). The Predictable scale had a significant interaction effect for type*acknowledgement ($F$(1,25)$ = 5.38$, $p = .029$, $\eta^2 = 0.18$). The Skilled scale showed a significant main effect of type ($F$(1,25)$ = 4.98$, $p = .035$, $\eta^2 = 0.17$). Finally, the Capable scale revealed a significant three-way interaction of type*timing*acknowledgement ($F$(1,25)$ = 6.99$, $p = .014$, $\eta^2 = 0.22$). The results showed that participants rated the robot higher on measures of perceived intelligence and trust in the questionnaire when the failure was executional, occurred early, or when the robot acknowledged its failure. However, for feelings of safety, ratings were higher when the failure occurred late and the robot did not acknowledge it. More information can be found in the Appendix.

\section{DISCUSSION}  


This study compared behavioural responses to robot failures, focusing on how individuals reacted and perceived the robot. Failures varied by type, timing, and acknowledgement. The findings revealed that robot failures affect user gaze and perceptions. These findings are discussed further in the following section.

\subsection{Behavioural Response}

To address the first research question, we analysed user gaze behaviour in multiple ways: the number of gaze shifts, gaze distribution during puzzle-solving, and gaze entropy based on transition matrices. These measures allowed us to examine how the type and timing of failures, as well as whether the robot acknowledged its failure, influenced user gaze patterns and whether gaze behaviour varied across different failure scenarios. Our results showed that user gaze is a reliable indicator of robot failures. When the robot made a failure, participants exhibited more frequent gaze shifts between different AoIs, likely due to confusion and an attempt to understand what was happening. This finding is similar to the results of Kontogiorgos et al. \cite{kontogiorgos_embodiment_2020}, who found that people tend to gaze more at the robot when it makes a mistake. The literature suggests that different types of failures influence user perceptions of the robot \cite{morales_interaction_2019}, and our findings support this by showing that users exhibit distinct gaze behaviours in response to various failure types. For example, when the failure was executional, the number of gaze shifts towards the robot was significantly higher compared to when the failure was decisional. Moreover, during executional failures, the proportion of time spent looking at the robot was much higher compared to decisional failures. It is crucial for the robot to recognize the type of failure it has made so that it can determine the appropriate strategy for recovery and regain the user's trust.

The timing of the failure is also crucial for the robot, as it requires different approaches for recovery and repair. In our research, while the timing of the failure—whether at the start or end of the interaction—did not significantly affect gaze shifts, it did influence gaze transition matrices, and gaze distribution across AoIs. Failures at the beginning of the interaction led to higher median gaze transition values, indicating more randomness early on. Additionally, participants' focus on the Tangram figure was more when the failure occurred at the beginning of the interaction compared to later ones, while their focus on the robot's body or end effector was more during late failures than early ones.


In our research, after committing a failure, the robot could either acknowledge the failure and then continue its action, or proceed without acknowledgement. We could not find significant differences in users' gaze behaviour when the robot acknowledged its failure and when it did not. As the literature suggests \cite{esterwood_you_2021, karli_what_2023, wachowiak_when_2024}, there are other verbal approaches to failure recovery, such as promises and technical explanations, which might influence users' gaze differently. Verbal failure recovery is important for robots, as it demonstrates an awareness of mistakes. This, in turn, can make the robot appear more intelligent and encourage users to engage with it more.

Our study also explored changes in users' anticipatory gaze behaviour during the task and its potential role in assisting the robot to recover from failures. Participants frequently anticipated the placement of the object before the robot executed the action, even when the robot made an error. This anticipatory gaze behaviour could serve as a valuable cue for the robot to detect its failures and initiate appropriate recovery strategies. However, we observed a decrease in participants' anticipatory gaze behaviour as the number of tasks increased. This decline may indicate reduced engagement over time, with participants being more actively collaborative at the beginning of the interaction. It also suggests that users' gaze behaviour might change throughout the interaction. These findings highlight the dynamic nature of gaze behaviour throughout the interaction.

\subsection{Subjective Measures}

To address the second research question, we examined user perceptions of the robot in three areas: perceived intelligence, sense of safety, and trust during failures. The analysis revealed how these measures varied with the type and timing of failure and whether the robot acknowledged its mistake.

The results of the subjective evaluation revealed that users' perceptions of the robot's intelligence and safety were not significantly influenced by the type of failure. However, users exhibited higher levels of trust in the robot during executional failures compared to decisional failures, suggesting that placing an object in an incorrect location reduces trust more than making an incorrect decision. Additionally, we observed interesting findings regarding the timing of the robot's failures. When failures occurred early in the interaction, users rated the robot as more intelligent and trustworthy compared to failures that occurred later. For the measure of "Sensible," this difference was statistically significant. These findings are consistent with previous research by Morales et al. \cite{morales_interaction_2019} and Lucas et al. \cite{lucas_getting_2018}. Interestingly, users reported feeling more relaxed when failures occurred later in the interaction, aligning with results from Desai et al. \cite{desai_impact_2013} and Rossi et al. \cite{rossi_how_2017}.

When the robot acknowledged its failures, users perceived it as slightly more intelligent and trustworthy but also experienced increased anxiety. This finding may be explained by the robot’s consistent physical repair actions a few seconds after each failure. When the robot did not explicitly acknowledge its failures, users might not have interpreted these actions as errors, reducing their perception of failure events.

\subsection{Limitations and Future Work}
 There were instances where participants were preoccupied with determining the placement of their next piece, which occasionally led them to overlook the robot's movements. However, these occurrences were minimal. Another limitation is the restriction to only two types of failure and whether the robot acknowledges its failure or not. The effect size in our study was medium; however, to obtain more robust results, a larger sample size would be beneficial.
 Furthermore, for safety reasons, the robot's arm movement was slowed and the experimenter was in the room, which may have influenced participants' perceptions. 
 Future research could address these limitations by exploring a broader range of failure types and incorporating explanatory feedback from the robot.

\section{CONCLUSION}

This study examines how robotic failures affect human gaze dynamics and perceptions during collaborative tasks, offering insights into using gaze as a failure indicator to assist in repair. The findings reveal that executional failures lead to more gaze shifts toward the robot, indicating user confusion, while decisional failures result in lower entropy in gaze transitions among areas of interest. 
Failures at the beginning of the interaction lead to more randomness in gaze shifts across AoIs. The timing of the failure during the task also affects users' gaze distribution across AoIs. Finally, acknowledgement of failure does not seem to affect gaze behaviour or users' perception. 
Our work contributes to a better understanding of how gaze behaviour can be leveraged in HRC to design more effective and reliable human-robot interaction systems.


\bibliographystyle{ieeetr}
\balance
\bibliography{references}

\clearpage
\onecolumn
\appendix\label{App}

Results of the three-way mixed ANOVA for subjective measures

\begin{table}[h!]
\centering
\resizebox{0.9\textwidth}{!}{
\begin{tabular}{c||c||c c c c c c c}

Scale  & Measure & Type & Timing & Acknowledgement & [Type*Timing] & [Type*Acknowledgement] & [Timing*Acknowledgement] & [Type*Timing*Acknowledgement] \\ \hline
Competent & df & $$(1,24)$$ & $(1,24)$ & $(1,24)$ & $(1,24)$ & $(1,24)$ & $(1,24)$ & $(1,24)$ \\ 
 & F value & 1.51 & 0.03 & 0.70 & 6.79 & 2.49 & 0.03 & 2.75 \\ 
& p value & .231 & .855 & .413 & \textbf{.016} & .127 & .855 & .110 \\ 
& $\eta^2$    & 0.06 & $<$0.01 & 0.03 & 0.22 & 0.09 & $<$0.01 & 0.10 \\ \hline
Sensible & df & $(1,24)$ & $(1,24)$ & $(1,24)$ & $(1,24)$ & $(1,24)$ & $(1,24)$ & $(1,24)$ \\ 
 & F value & 2.63 & 5.79 & 0.53 & 0.44 & 2.62 & 1.79 & 2.41 \\ 
& p value & .118 & \textbf{.024} & .475 & .512 & .118 & .194 & .134 \\ 
& $\eta^2$    & 0.10 & 0.19 & 0.02 & 0.02 & 0.10 & 0.07 & 0.09 \\  \hline
Responsible & df & $(1,24)$ & $(1,24)$ & $(1,24)$ & $(1,24)$ & $(1,24)$ & $(1,24)$ & $(1,24)$ \\ 
 & F value & 0.58 & 0.04 & 1.74 & 0.03 & 0.21 & 0.34 & 0.29 \\ 
& p value & .453 & .848 & .200 & .859 & .651 & .567 & .595 \\ 
& $\eta^2$    & 0.02 & $<$0.01 & 0.07 & $<$0.01 & $<$0.01 & 0.01 & 0.01 \\  \hline
Anxious/Relaxed & df & $(1,24)$ & $(1,24)$ & $(1,24)$ & $(1,24)$ & $(1,24)$ & $(1,24)$ & $(1,24)$ \\ 
(Self) & F value & 0.07 & 7.80 & 2.23 & 0.26 & 5.50 & 0.84 & 0.01 \\ 
& p value & .792 & \textbf{.010} & .148 & .613 & \textbf{.027} & .369 & .905 \\ 
& $\eta^2$  & $<$0.01 & 0.24 & 0.08 & 0.01 & 0.18 & 0.03 & $<$0.001 \\ \hline

Reliable & df & $(1,25)$ & $(1,25)$ & $(1,25)$ & $(1,25)$ & $(1,25)$ & $(1,25)$ & $(1,25)$ \\ 
 & F value & 1.61 & 0.83 & 1.28 & 1.50 & 1.00 & 0.35 & 0.99 \\ 
& p value & .216 & .370 & .268 & .233 & .328 & .561 & .330 \\ 
& $\eta^2$   & 0.06 & 0.03 & 0.05 & 0.06 & 0.04 & 0.01 & 0.04 \\  \hline

Predictable & df & $(1,25)$ & $(1,25)$ & $(1,25)$ & $(1,25)$ & $(1,25)$ & $(1,25)$ & $(1,25)$ \\ 
& F value & 2.98 & 1.43 & 0.94 & 3.97 & 5.38 & 0.26 & 3.97 \\ 
& p value & .097 & .243 & .340 & .057 & \textbf{.029} & .616 & .057 \\ 
& $\eta^2$  & 0.11 & 0.05 & 0.04 & 0.14 & 0.18 & 0.01 & 0.14 \\ \hline

Skilled  & df & $(1,25)$ & $(1,25)$ & $(1,25)$ & $(1,25)$ & $(1,25)$ & $(1,25)$ & $(1,25)$ \\ 
& F value & 4.98 & 2.93 & 0.43 & 0.13 & 1.65 & 0.11 & 3.46 \\ 
& p value & \textbf{.035} & .099 & .516 & .719 & .210 & .741 & .075 \\ 
& $\eta^2$    & 0.17 & 0.11 & 0.02 & $<$0.01 & 0.06 & $<$0.01 & 0.12 \\  \hline

Capable & df & $(1,25)$ & $(1,25)$ & $(1,25)$ & $(1,25)$ & $(1,25)$ & $(1,25)$ & $(1,25)$ \\ 
& F value & 1.71 & 3.02 & 0.80 & 1.15 & 0.002 & 0.41 & 6.99 \\ 
& p value & .203 & .095 & .380 & .293 & .962 & .528 & \textbf{.014} \\ 
& $\eta^2$    & 0.06 & 0.11 & 0.03 & 0.04 & $<$0.001 & 0.02 & 0.22 \\ \hline

\end{tabular}
}
\label{subjective}
\end{table}

\end{document}